\documentclass[conference]{IEEEtran}
\IEEEoverridecommandlockouts

\usepackage{cite}
\usepackage{amsmath,amssymb,amsfonts}
\usepackage{algorithmic}
\usepackage{graphicx}
\usepackage{textcomp}
\usepackage{xcolor}
\usepackage{booktabs}
\usepackage{multirow}

\def\BibTeX{{\rm B\kern-.05em{\sc i\kern-.025em b}\kern-.08em
    T\kern-.1667em\lower.7ex\hbox{E}\kern-.125emX}}

\begin{document}

\title{QMAVIS: Long Video-Audio Understanding using Fusion of Large Multimodal Models\\

}

\author{
\IEEEauthorblockN{
Zixing Lin\textsuperscript{1}, 
Jiale Wang\textsuperscript{1}, 
Gee Wah Ng\textsuperscript{1}, 
Lee Onn Mak\textsuperscript{1}, \\
Chan Zhi Yang Jeriel\textsuperscript{2}*, 
Jun Yang Lee\textsuperscript{3}*, 
Yaohao Li\textsuperscript{3}*
}

\IEEEauthorblockA{\vspace{0.5em}}

\IEEEauthorblockA{\textsuperscript{1}Q Team, Home Team Science and Technology Agency (HTX), Singapore
}
\IEEEauthorblockA{\textsuperscript{2}National University of Singapore}
\IEEEauthorblockA{\textsuperscript{3}Nanyang Technological University, Singapore}

\thanks{* Research conducted during an internship at Q Team, HTX.}
}

\maketitle

\begin{abstract}
Large Multimodal Models (LMMs) for video-audio understanding have traditionally been evaluated only on shorter videos of a few minutes long. In this paper, we introduce QMAVIS (\textbf{Q} Team-\textbf{M}ultimodal \textbf{A}udio \textbf{V}ideo \textbf{I}ntelligent \textbf{S}ensemaking), a novel long video-audio understanding pipeline built through a late fusion of LMMs, Large Language Models, and speech recognition models. QMAVIS addresses the gap in long-form video analytics, particularly for longer videos of a few minutes to beyond an hour long, opening up new potential applications in sensemaking, video content analysis, embodied AI, etc. Quantitative experiments using QMAVIS demonstrated a \textbf{38.75\% improvement }over state-of-the-art video-audio LMMs like VideoLlaMA2 and InternVL2 on the VideoMME (with subtitles) dataset, which comprises long videos with audio information. Evaluations on other challenging video understanding datasets like PerceptionTest and EgoSchema saw up to 2\% improvement, indicating competitive performance. Qualitative experiments also showed that QMAVIS is able to extract the nuances of different scenes in a long video audio content while understanding the overarching narrative. Ablation studies were also conducted to ascertain the impact of each component in the fusion pipeline. 
 
\end{abstract}


\section{Introduction}

Early Large Language Models (LLM) such as GPT-3 \cite{gpt3_paper} and LLaMA \cite{llama2_paper} kicked off a wave of AI innovations, thanks to their remarkable ability to process natural language prompts and generate coherent responses. Riding on the AI wave, Large Multimodal Models (LMM) emerged, integrating other modalities beyond text, such as vision. Models such as GPT-4 \cite{gpt4_paper} and LLaVA \cite{llava_paper} fused language and visual understanding, enabling AI to engage with the world more completely through images in addition to text. The next advancement involved LMMs for video comprehension. Models and pipelines such as Video-LLaVA \cite{videollava_paper}, Video-ChatGPT \cite{maaz2023videochatgpt}, and QCaption \cite{QCaption} extended AI's capabilities further by incorporating temporal reasoning and sequential frame processing, opening up new possibilities in video captioning and Q\&A, taking AI a step closer to human cognition. However, a crucial aspect of human perception had been largely overlooked in prior research: audio.

Multimodal models that jointly comprehend video and audio, or video-audio LMMs, is an emerging field of research that has become increasingly crucial for AI systems to understand the world. Just like how a human would watch a video or explore the world around him, taking in both audio and visual information and responding to it, the goal of these AI models has been to align visual events with auditory cues, on top of the usual textual information, painting a more complete picture of the scenario. Notable trailblazing models include Video-LLaMA \cite{VideoLlama} and PandaGPT \cite{PandaGPT}. By fusing textual, visual, and audio cues, Video-audio LMMs provide a more holistic perception of the world around it, and could open ample applications in domains like multimedia content analysis (e.g., movies), sensemaking for public safety, autonomous vehicles, and embodied AI.

Despite recent progress in video-audio LMMs,\textbf{ \textbf{there remains a gap in processing longer videos (e.g.,  a few minutes to beyond an hour long)}, while comprehending both visual and audio information comprehensively} due to architectural and computational limitations \cite{VideoXL}. On the architectural front, existing video LMMs, regardless of whether they include audio as an additional modality, often down-samples the video significantly, extracting only a small number of (still) frames and audio clips from the entire video. This leads to increased probability of missed temporal context and piecewise understanding of long content, particularly for videos with multiple scene changes \cite{QCaption}. On the computational front, thousands of visual tokens for long videos exceeds typical LMM context windows, resulting in an inability to analyse these comprehensively at one go \cite{VideoXL}. This unaddressed gap motivates QMAVIS, which specifically targets long video-audio understanding by using a late fusion of model outputs to process videos in manageable chunks, thus maintaining temporal information as compared to static frame snapshots while concurrently circumventing context window limitations. 

\subsection{Key Contributions}

We present the \textbf{QMAVIS} (\textbf{Q} Team-\textbf{M}ultimodal \textbf{A}udio \textbf{V}ideo \textbf{I}ntelligent \textbf{S}ensemaking) pipeline which adopts a late fusion strategy to enable processing longer video and audio context, by chunking the inputs, analysing them in separate LLMs, video LMMs, and audio-text models before fusing their outputs together to form a coherent response conditioned on the input prompt, thereby preserving rich multi-modal context for longer videos. The design of QMAVIS is also modular: models can be swapped out for future more capable models, or models finetuned for specific use cases. The entire QMAVIS pipeline can be run fully on-premises and using open-source models, without reliance on external APIs, an important requirement for use cases requiring data privacy.

Through experiments on standardized Video Understanding benchmarks, we demonstrated the prowess of QMAVIS in processing long-form video-audio information. \textbf{Notably, QMAVIS outperformed leading baseline video LMMs like VideoLlama2 \cite{VideoLlama} and InternVideo2 \cite{InternVideo2} by \textbf{38.75\%} on the VideoMME dataset, of which more than half are comprised of long videos between 4 mins to beyond an hour long. }On PerceptionTest, which comprises rich and complex video analytics tasks, QMAVIS outperformed baselines by up to 2\%.

We also conducted ablation studies to assess the impact of different fusion components, for example the response aggregation LLM, on the overall results.

\section{Related Works and Fusion techniques}

\subsection{Video LMM}

Video LMMs such as Video-LLaVA \cite{videollava_paper} and Video-ChatGPT \cite{maaz2023videochatgpt} are able to process complex video content, interpret prompts, and generate coherent responses. This enables tasks like video captioning and natural language-based video Q\&A. However, these models perform best on short clips ranging from a few seconds to 1–2 minutes and struggle with multi-scene videos and capturing temporal relationships \cite{videollava_paper}. QCaption used a late fusion approach to integrate chunked outputs from LLMs and LMMs to analyse longer videos, achieving improvements of up to 44.2\% in video captioning and 48.9\% in Q\&A tasks for longer videos compared to models like Video-LLaVA \cite{QCaption}. However, these works have been limited to analysing videos only, without the ability to process audio as an additional modality.

\subsection{Video-audio LMM}

Recent Video-audio LMMs have explored different fusion strategies to combine video, audio, and textual context for comprehensive scene analysis. \textbf{VideoLLaMA} \cite{VideoLlama} employs an intermediate to late fusion approach, aligning vision and audio features in the language domain through learned query transformers. Specifically, video and audio Q-Formers are respectively used to encode video frames and audio signals, before projection into the LLM's embedding space to generate the final answer, conditioned on the input prompt. \textbf{InternVideo2} \cite{InternVideo2}, which is often cited as a state-of-the-art video-audio model, takes a late fusion approach to integrate video and text embeddings for the LLM to generate responses. The paper presented a novel approach emphasising spatiotemporal consistency – during training it generates video-audio-speech captions to ensure that visual and auditory information are jointly grounded in text, ensuring better alignment. Similarly, \textbf{PandaGPT} \cite{PandaGPT} also presents a late-fusion approach, leveraging ImageBind multi-modal embedding model to produce common embeddings for images and audio, and feeding them into an LLM. Notably, this approach allowed the model to exhibit emergent audio-visual understanding, despite only being trained on image-text pairs (i.e., without explicit audio inputs). Early fusion techniques, such as merging audio and visual inputs at input time are not commonly used in recent video-audio LMMs. Instead, variants of intermediate to late fusion dominate, mainly through late-stage fusion via shared embedding spaces or common modalities (e.g., text). Irregardless of the fusion approach taken, prior work all centered around training models for handling short video clips in the order of 10-30 seconds \cite{VideoLlama, InternVideo2, videollava_paper}, without an explicit focus on longer videos from a few minutes to beyond an hour long.

\section{Methodology}

\begin{figure*}[ht!]
  \centering
  \includegraphics[width=\textwidth]{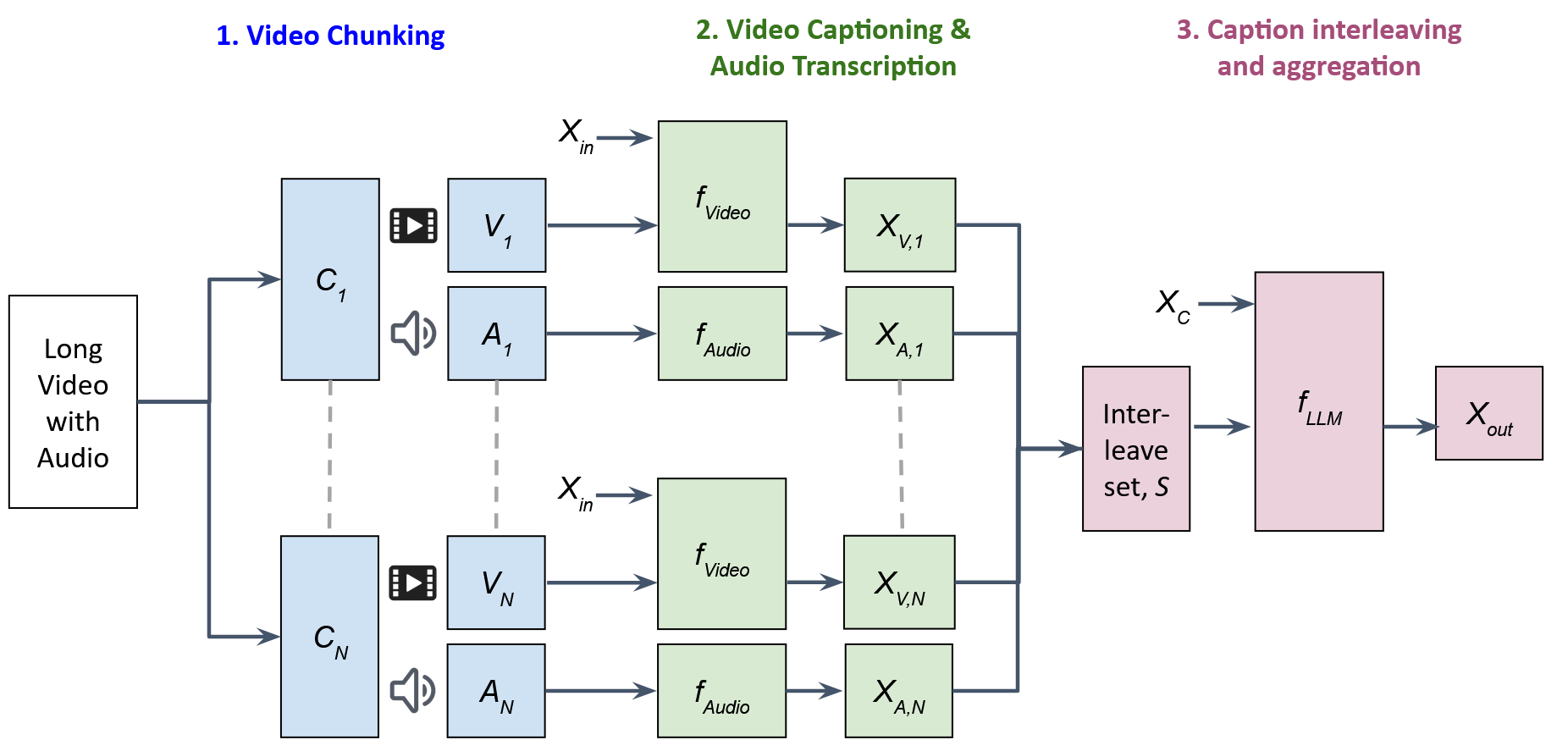}
  \caption{QMAVIS Fusion Architecture.}
  \label{fig:qmavis_arch}
\end{figure*}

\subsection{Fusion Architecture}

To achieve effective long video captioning with audio context, QMAVIS fuses the outputs of three large models in a single pipeline: Video captioning LMM, an audio transcription model, and an LLM (Figure \ref{fig:qmavis_arch}).

\subsection{Video Chunking}

Starting from the input video that contains an audio track, the video is first split into \( N \) short chunks, each of about 30 seconds long, \(C_i, \text{for } i \in \{1, 2, \ldots, N\}\); the longer the video, the more video chunks. Every frame of the video clip will be included in one of the chunks to ensure that no information is left out during the analysis. From each chunk, the visual information (the video clip), \(V_i, \text{for } i \in \{1, 2, \ldots, N\}\);  and audio information (the audio file), \(A_i, \text{for } i \in \{1, 2, \ldots, N\}\), will be separately extracted and indexed for further analysis.

\subsection{Video Captioning and Audio Transcription}

In this stage, for each chunk \( C_i \), the visual  \( V_i \) and audio  \( A_i \) information are separately processed. 

The video clip,  \( V_i \), is fed into a Video LMM, \(f_{video}\), along with an input prompt  \(X_{in}\), to generate a caption or answer conditioned on the prompt, \( X_{V,i} \), indicated by equation \ref{eq1}. The input prompt,  \(X_{in}\), are kept consistent for each benchmarking experiment but varies across experiments depending on the dataset used. For example, for qualitative experiments,  \(X_{in}\) is simply "Describe this video in detail", while for PerceptionTest,  \(X_{in}\) is "Taking into account the following question: \texttt{\{question\}}, describe this video in detail", owing to the varying input prompts posed under the dataset. The Video LMM used for QMAVIS experiments and ablations are standardised to Qwen2-VL-72B \cite{Qwen}.

\begin{equation}
X_{v,i} = f_{\text{video}}(X_{\text{in}}, V_i), \quad \text{for } i \in \{1, 2, \ldots, N\}
\label{eq1}
\end{equation}

The audio clip, \( A_i \), is transcribed to text using Whisper Large v3 \cite{Whisper}, a Speech Translation / Automatic Speech Recognition (ASR) model,  \(f_{audio}\). Unlike the video captioning task, the audio transcription output,  \( X_{A,i} \).  is not conditioned on an input prompt, but a simple one-to-one translation of audio to text, indicated by equation \ref{eq2}. In this process, the audio waveform of the chunk first transformed into a spectrogram, then passed through a sequence-to-sequence model trained on vast datasets of multilingual speech corpus across a range of tasks, including multi-speaker diarisation, yielding a good transcription of the audio clip for further analysis.

\begin{equation}
X_{A,i} = f_{\text{audio}}(A_i), \quad \text{for } i \in \{1, 2, \ldots, N\}
\label{eq2}
\end{equation}

\subsection{Caption Interleaving and Aggregation}

After each individual chunk has been captioned and transcribed, we then assemble a set  \(\mathcal{S}\), which comprises the video caption of each chunk, \( X_{V,i} \), interleaved with its associated audio transcription, \( X_{V,i} \).  That way, we ensure that each chunk is analysed holistically with both visual and audio information, considered. Equation \ref{eq3} illustrates the interleaving operation.

\begin{equation}
\mathcal{S} = \bigcup_{i=1}^{N} \{ X_{V,i}, X_{A,i} \}
\label{eq3}
\end{equation}

The entire set  \(\mathcal{S}\) is then passed through an LLM to produce an aggregated textual response with reference to information gleaned from the separate chunks. The response is conditioned on a second input prompt,  \( X_{C} \),  which tasks the LLM to consider the individual chunks of information and aggregate them into a complete, coherent report while paying attention to the nuances of different scenes that is portrayed in the long video. \( X_{C} \) is kept consistent across all experiments irregardless of the dataset. Where the number of tokens in the set exceeds the LLM context window, these captions are then aggregated recursively.  

\subsection{Ablation Studies}

To investigate the effects of each component of the fusion architecture, we conducted ablation experiments by disabling select components systematically. 

One experiment involved disabling the caption aggregation LLM, and presenting the entire set  \(\mathcal{S}\)  as a naive concatenation of textual outputs as the final answer without further summarization or processing. The goal is to investigate the tradeoffs between potential details lost in the captioning aggregation stage, against a more succinct and coherent response.

Another experiment involved disabling the audio transcription model, to assess the impact of the loss of audio information.

\section{Results}

\subsection{Evaluation Metrics}

Evaluating multi-modal models requires a quantitative metric to assess their ability to interpret, reason and understand across different modalities, including video, audio, and text. One of the most widely used evaluation methods for multiple-choice question-answering (QA) tasks is Top-1 accuracy.

This metric is robust in zero-shot evaluations of multi-modal models across benchmarks such as the Perception Test mc-VQA, VideoMME, and EgoSchema. These benchmarks test various aspects of multi-modal understanding, including general comprehension, context recognition, and task-specific reasoning. By assessing the correctness of the model’s predictions, Top-1 accuracy provides an effective comparison metric for evaluating models.

In this evaluation, models are required to answer multiple-choice questions derived from real-world video and audio data. The task is to determine how accurately a model can process complex multi-modal inputs and select the correct response from a set of choices.

Top-1 accuracy is computed using the following formula:  

\begin{equation}
    \text{Top-1 Accuracy} = \left( \frac{\text{Number of Correct Predictions}}{\text{Total Number of Questions}} \right) \times 100
\end{equation}

To further analyze model performance, confidence intervals can be computed to quantify the statistical significance of the reported accuracy

\begin{equation}
    CI = \hat{p} \pm Z \sqrt{\frac{\hat{p} (1 - \hat{p})}{N}}
\end{equation}  

where \( \hat{p} \) is the observed Top-1 accuracy, \( N \) is the total number of questions, and \( Z \) is the critical value corresponding to a given confidence level (e.g., 1.96 for 95\% confidence).

\subsection{Benchmark Datasets}

We delve into the three benchmark datasets used:  \textit{Perception Test mc-VQA}, \textit{VideoMME}, and \textit{EgoSchema}.

\subsubsection{Perception Test}  

\subsubsection{VideoMME}  

\textit{VideoMME}, by Chaoyou Fu et al. \cite{fu2024video}, focuses on evaluating multi-modal models in video-based question-answering (QA) and captioning tasks. It features diverse videos across six domains and 30 subfields. A key feature of VideoMME is its emphasis on videos of varying durations, ranging from as short as 11 seconds to over 1 hour. For the VideoMME set used in our experiments, \textbf{52\% of the videos are longer videos between 4 mins to over an hour long}, while the remaining 48\% of videos are shorter videos from 11 seconds to 4 mins long.  

VideoMME uses multi-modal inputs, including video, subtitles, and audio, ensuring a comprehensive assessment. High-quality annotations provide precise evaluation. The evaluation is designed to measure a model's ability to retain short-term contextual knowledge, identify trends and patterns in medium-length videos, and infer conclusions from long-term sequences. Apart from its emphasis on video length, VideoMME uses multiple data modalities, including subtitles and audio, in addition to video frames. This enables a comprehensive assessment of the multimodal reasoning capabilities of a model.

\subsubsection{EgoSchema}  

The EgoSchema benchmark, created by Karttikeya Mangalam et al. \\\cite{EgoSchema}, aims to test a multi-modal model’s ability to understand and reason about long-term events in ego-centric video data. Unlike other datasets like Perception Test \cite{perceptiontest2023} that focus on short-term visual perception, EgoSchema offers a longer set of video clips, with a median length of approximately 100 seconds. The dataset includes a wide range of real-world scenarios, from everyday activities to highly task-specific environments, challenging models to maintain contextual awareness over time. The benchmark's ability to assess different levels of temporal reasoning is one of its key features. Models are tested on their recognition of short-term activities, such as playing an instrument or taking notes, as well as more complex, medium-term tasks, such as organizing a workspace or tracking a conversation over time. Long-term reasoning is also required, as models must abstract overarching behavioural patterns from extended video sequences, identifying themes and objectives that emerge over time. This benchmark offers a challenging assessment of the multi-modal model for long-term contextual reasoning by pushing models to find connections between activities that are not immediately consecutive.

\subsection{Qualitative Experiments}

We also performed qualitative experiments on videos over an hour long, often featuring multiple scenes and an overarching narrative/storyline to follow. The goal is to compare the ability of QMAVIS, vis-a-vis baseline models, to capture nuances in each scene while keeping track of the overall narrative.

\subsection{Baseline}

We selected three well-known video-audio LMMs as the baseline to validate the QMAVIS fusion pipeline. 

\textbf{InternVideo2} \cite{InternVideo2} is a state-of-the-art video-audio foundation model that adopts a late fusion approach to ensure that visual and auditory information are jointly grounded in the text, ensuring better alignment. \textbf{PandaGPT} \cite{PandaGPT} is a multi-modal instruction-following model capable of processing both images and audio. It employed late fusion by generating common embeddings for image and audio inputs, which are then fed into an LLM via a projection layer. \textbf{VideoLLaMA2} \cite{VideoLlama} is another well known video-audio LMM that employs late fusion by first encoding visual and audio features separately and then aligning them within a shared embedding space before ingesting them to the LLM for reasoning and generation. Each video-audio LMM, alongisde QMAVIS, were evaluated on the three above-mentioned benchmark datasets: VideoMME, PerceptionTest, and EgoSchema.  Top-1 accuracy scores were reported for all three datasets.
\begin{figure*}[ht!]
  \centering
  \includegraphics[width=0.8\textwidth]{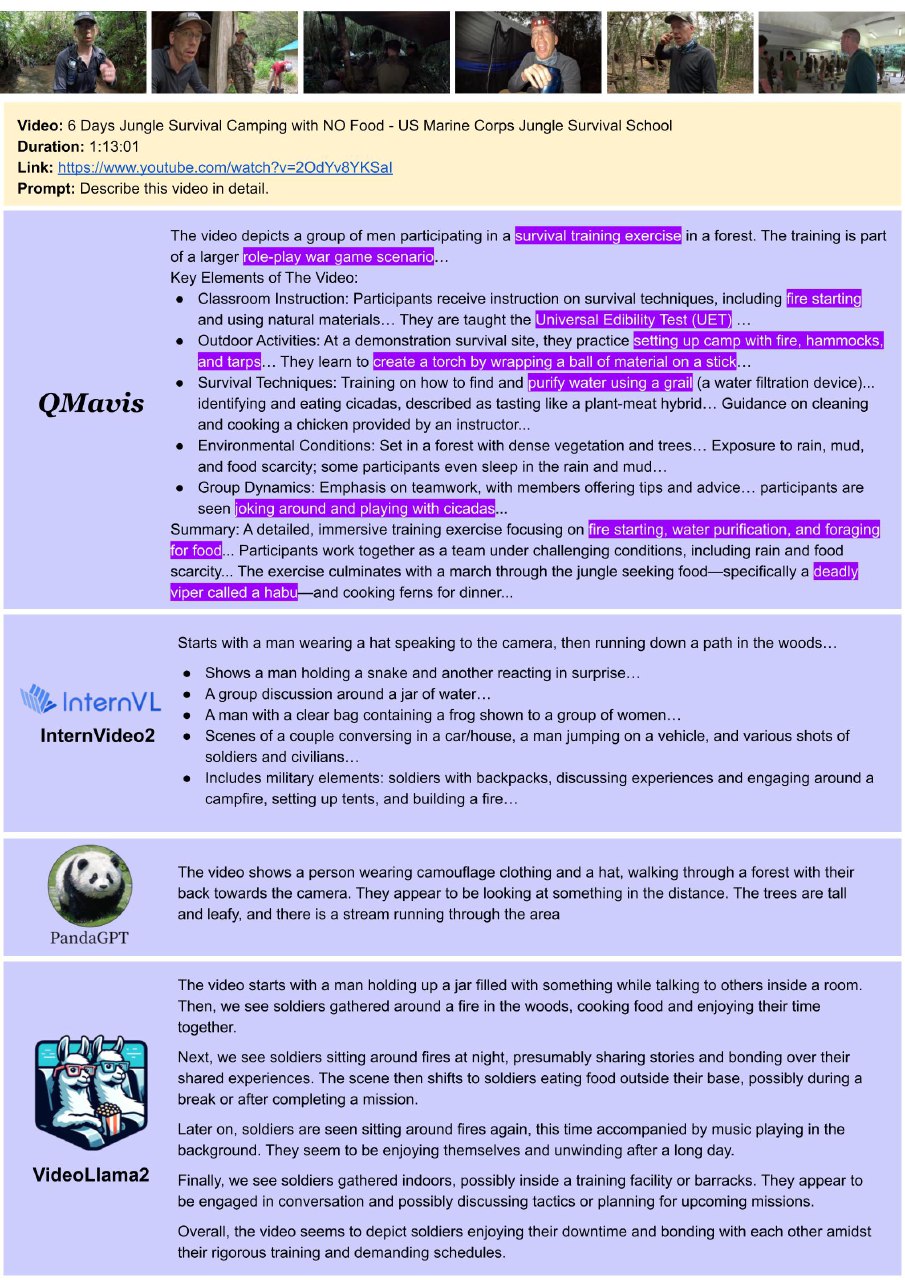}
  \caption{Qualitative Results between QMAVIS, InternVideo2 PandaGPT and VideoLLaMA2}
  \label{fig:qmavis_qual}
\end{figure*}
\subsection{QMAVIS}

\begin{table*}[htbp]
\small 
\setlength{\tabcolsep}{3pt} 
\centering
\caption{Performance comparison between QMAVIS and other video understanding models across multiple benchmarks.}
\label{tab:model_comparison}
\begin{tabular}{clp{3cm}p{3cm}p{2cm}ccc}
\toprule
\multirow{2}{*}{S/N} & \multirow{2}{*}{} & \multirow{2}{*}{\textbf{Method}} & \multirow{2}{*}{\textbf{Video LMM used}} & \multirow{2}{*}{\textbf{Audio}} & \multicolumn{3}{c}{\textbf{Video Understanding Benchmarks}} \\
\cmidrule(lr){6-8}
& & & & & VideoMME (w/ subs)  & Perception Test & EgoSchema \\
\midrule
1 & \multirow{3}{*}{Baseline} & VideoLLaMA2 & VideoLLaMA2 & Included in VideoLLaMA2 & 47.9 & 57.5 & 63.9 \\
2 & & InternVideo2 (Base) & InternVideo2 & Included in InternVideo2 & 41.9 & 52.16 & 55.0 \\
3 & & PandaGPT & ImageBind & Included in PandaGPT & 22.58 & 31.63 & 24.0 \\
\midrule
4 & QMAVIS & QMAVIS (full) & Qwen2VL & Whisper-LargeV3 & \textbf{66.46}* & 57.72 & \textbf{65.0} \\
\midrule
5 & \multirow{2}{*}{Ablation Studies} & QMAVIS (Ablation 1 - no LLM) & Qwen2VL & Whisper-LargeV3 & {63.00}* & 53.14 & 62.8 \\
6 & & QMAVIS (Ablation 2 - no STT) & Qwen2VL & None & 48.18* & \textbf{58.37} & - \\
\bottomrule
\end{tabular}
\vspace{1em}
\begin{minipage}{\textwidth}
\footnotesize
* Partial dataset (\~75\% of total dataset), experiments still ongoing.
\end{minipage}
\end{table*}

Referring to Table \ref{tab:model_comparison}, with a score of 66.46 on VideoMME (with subtitles), QMAVIS demonstrated a \textbf{38.75\%} improvement over the next best performing model, VideoLlama2. 52\% of the videos in VideoMME are between 4 mins to over an hour long, among which about 15\% are long videos that are above 15 mins in duration. All videos have corresponding audio tracks. This demonstrates the benefits of using a fusion approach to analyse long videos, by breaking them down into shorter chunks that current video LMMs excel in and leveraging the aggregation abilities of an LLM to produce a final coherent response. We performed experiments on the different video lengths using parameters optimized for the video duration. Video chunks of 60 seconds with a 1 frame per second (FPS) sampling rate were used for short and medium videos, while 600 seconds with 0.5 FPS was used for long videos for a more thorough and less context-heavy aggregation in the final step. We observe that the chunking approach allow for powerful long video understanding, preserving details that will otherwise be lost when ingesting the entire video.

Likewise, on PerceptionTest, which focuses on robust and complex situations, QMAVIS without the audio component attained a \textbf{1.51\%} improvement over VideoLLaMA2. While the difference is small, the approach still demonstrates QMAVIS' ability to surpass state-of-the-art models in challenging video understanding tasks. The chunking approach presents a method to focus the video LMM on small sections of the video, analyzing the details of complex scenarios at a high sampling rate. The smaller chunked descriptions are passed to the fusion LLM, which acts as an important component in memory-related tasks, allowing for greater context understanding in text space. 

We also observe improvements in EgoSchema, a dataset that measures a model's performance in real-world scenarios and both short and long-term contextual reasoning. QMAVIS surpasses the performance of VideoLlama by \textbf{1.72\%}, which demonstrates the ability of aggregation LLM to draw temporal links between video descriptions from each chunk, as well as the concise descriptions of the chunks from the video LMM.

In the three benchmarks, we observe the greatest boost in performance in VideoMME as compared to PerceptionTest and EgoSchema. This is likely attributed to QMAVIS' strengths in long video understanding, as compared to shorter videos. Despite having longer videos than an average video understanding dataset, PerceptionTest and EgoSchema still both primarily comprise videos that are around 22 seconds and 100 seconds long respectively, while VideoMME contain a greater volume of significantly longer videos of 4 minutes to beyond 1 hour each. With longer videos, the capabilities of the chunking and aggregation pipeline is more prominent which leads to overall superior results compared to other models.

\subsection{Ablation Results}

We also performed an ablation study of QMAVIS by disabling certain components in the pipeline. The goal is to investigate the tradeoffs between potential details lost in the captioning aggregation stage, against a more succinct and coherent response. 

For VideoMME (with subtitles), we first experimented with using the video LMM for aggregation, without a dedicated LLM. We observe a \textbf{3.72\%} drop in performance over the full QMAVIS, but maintain a \textbf{38.75\%} lead over VideoLLaMA2. This demonstrates the powerful reasoning ability of the video LMM in handling large textual contexts together with added audio information. While being inferior to a dedicated aggregation LLM, this presents a reliable alternate implementation for lighter-weight deployment where the number of models used can be reduced, with palatable declines in performance. We also experimented with removing the audio transcription model, which presents a significant drop in performance from the full model at \textbf{48.18\%} accuracy. This loss of support for the audio modality reduces QMAVIS to a pure visual implementation, which results in poor performance in all audio-related tasks. However, this approach maintains a slight lead over VideoLLaMA2 by \textbf{0.28\%}.

The same ablation studies were performed for PerceptionTest and EgoSchema. For both benchmarks, we observe a similar behaviour as VideoMME in the results without the LLM for aggregation. In the second ablation, however, the results for PerceptionTest demonstrated improvements in performance over the full QMAVIS pipeline, demonstrating a \textbf{0.65\%} boost in accuracy. As much of PerceptionTest's audio tasks involves non-spoken audio cues rather than pure speech, the removal of the audio transcription component likely reduced the inaccuracies introduced by potential misinterpretation of non-spoken audio signals. In the case of PerceptionTest, it was likely that the audio transcription model could have hindered the overall pipeline.

\subsection{Qualitative Results}

Figure \ref{fig:qmavis_qual} compares the responses generated by QMAVIS with respect to benchmark models on long videos (over an hour) with accompanying audio. The videos were chosen for their multiple distinct scenes and an overarching narrative.

InternVideo2:
Provides highly detailed descriptions of characters—down to what they are wearing and doing—but it fails to identify the multiple scenes or convey the overall storyline.

PandaGPT:
Offers a concise, single-frame description that captures one scene well but misses the temporal and narrative context necessary for long videos.

VideoLlama2:
Presents a narrative-focused summary that covers group interactions and sequential events; however, it does not deliver scene-level details with sufficient nuance.

QMAVIS:
Excels by extracting the key scenes in the video and providing detailed descriptions for each, followed by a comprehensive overview of the overall flow of events. This approach allows QMAVIS to capture both fine-grained details and the broader narrative context.

Overall, QMAVIS outperforms the benchmark models by effectively balancing granular detail with an understanding of the narrative structure in long, complex videos.

\section{Conclusion}

In this paper, we presented QMAVIS, a novel audio-video captioning and Q\&A pipeline that adopts a late fusion approach to effectively process long-form video and audio content (up to beyond an hour long), capturing the overarching narrative while retaining the nuances from scenes within the video. This addresses a gap in video-audio LMMs, which are often limited to short clips in the order of 10-30 seconds long.  QMAVIS outperformed leading baseline models, such as VideoLlama2 and InternVideo2, on Video Understanding benchmarks VideoMME, PerceptionTest, and EgoSchema, by \textbf{38.75\%}, 1.51\%, 1.72\% respectively. Notably, QMAVIS' capabilities on long-form video-audio understanding are apparent from the large improvement in the VideoMME dataset, of which more than half are comprised of longer videos from 4 mins to beyond an hour long.

\textbf{Future work.} There is potential to expand the scope of study to audio models that not just transcribes speech, but also comprehend tone and non-speech signals like background noise, music, and non-spoken cues (e.g., claps, footsteps), giving a much richer analysis of the video. One may also wish to explore other possibilities for the caption interleaving stage; instead of interleaving audio and video information by each chunk, perhaps the interleaving can be extended to multiple adjacent chunks, to potentially improve temporal understanding. There is also potential to tune certain hyperparameters in the fusion pipeline, for example the chunk length, number of chunks, frame sampling rate in video captioning, or context length of aggregation LLM, to yield more computationally efficient results. There is also potential to develop dedicated long-form video-audio evaluation datasets, as most existing datasets, including those used in this paper, still lack a wider library of longer videos beyond just a few minutes. Finally, as with most Generative AI-based pipelines, there is always room to further refine the input prompts to yield more precise answers or answers more tailored for specific use cases.

In all, we hope that QMAVIS lays a foundation for further research on robust audio-video LMM pipelines that employs a fusion approach. 

\section*{Acknowledgment}

This work is funded by the Home Team Science and Technology Agency (HTX), a statutory board under the Ministry of Home Affairs (MHA), Singapore.


\bibliographystyle{unsrt}
\bibliography{bibtex.bib}

\end{document}